\documentclass[letterpaper, 10 pt, conference]{ieeeconf}

\IEEEoverridecommandlockouts                           
\usepackage[T1]{fontenc}
\usepackage{amsmath,amsfonts}
\usepackage{algorithm, algpseudocode}
\usepackage{graphicx}
\usepackage{caption}
\usepackage{booktabs}
\usepackage{microtype}
\usepackage{bm}

\usepackage{tikz}
\usetikzlibrary{automata,positioning,arrows.meta,calc,decorations.markings}
\usetikzlibrary{fit,backgrounds}

\algdef{SE}[DOWHILE]{Do}{doWhile}{\algorithmicdo}[1]{\algorithmicwhile\ #1}%

\title{\LARGE \bf
    PAAMP: Polytopic Action-Set And Motion Planning for Long Horizon Dynamic Motion Planning via Mixed Integer Linear Programming
}

\author{Akshay Jaitly and Siavash Farzan%
\thanks{Akshay Jaitly is with the Robotics Engineering Department, Worcester Polytechnic Institute, 
Worcester, MA 01609, USA, {\tt\small ajaitly@wpi.edu}}%
\thanks{Siavash Farzan is with the Electrical Engineering Department, California Polytechnic State University, San Luis Obispo, CA 93407, USA, {\tt\small sfarzan@calpoly.edu}}
}

\begin{document}

\maketitle
\thispagestyle{empty}
\pagestyle{empty}

\begin{abstract}
Optimization methods for long-horizon, dynamically feasible motion planning in robotics tackle challenging non-convex and discontinuous optimization problems. Traditional methods often falter due to the nonlinear characteristics of these problems. We introduce a technique that utilizes learned representations of the system, known as Polytopic Action Sets, to efficiently compute long-horizon trajectories. By employing a suitable sequence of Polytopic Action Sets, we transform the long-horizon dynamically feasible motion planning problem into a Linear Program. This reformulation enables us to address motion planning as a Mixed Integer Linear Program (MILP). We demonstrate the effectiveness of a Polytopic Action-Set and Motion Planning (PAAMP) approach by identifying swing-up motions for a torque-constrained pendulum as fast as 0.75 milliseconds. This approach is well-suited for solving complex motion planning and long-horizon Constraint Satisfaction Problems (CSPs) in dynamic and underactuated systems such as legged and aerial robots.
\end{abstract}

\section{Introduction}

In the field of long-horizon, dynamically feasible motion planning for robotics, the optimization process involves solving non-convex and discontinuous optimization problems.
Traditional approaches generally struggle with the nonlinear nature of this problem, leading to solutions that are computationally intensive and not always practical for real-world applications. Certain desired behaviors may be completely infeasible to solve for without appropriate initial guesses~\cite{quadruped_jumping}. This paper presents a method that employs learned representations of the system to simplify the problem, representing it as a Mixed Integer Linear Program (MILP). The solutions to this MILP are guaranteed to be feasible trajectories.\looseness=-1

Our approach is comparable to optimization methods for dynamically feasible motion planning, such as direct transcription methods. These methods frame the problem of motion planning as a Nonlinear Program~\cite{direct-collocation}. Direct Single Shooting methods attempt to determine control inputs that, when simulated forward in time, satisfy specific boundary conditions. The consideration of dynamics over a long horizon results in a highly nonlinear relationship between the control parameter inputs and the simulated final state.

Parallel methods, like Multiple Shooting and Direct Collocation, attempt to solve the same problem by instead considering the trajectory as compositions of multiple segments. The initial state of each trajectory segment ($x_{t_0}, x_{t_1}, x_{t_2}, \ldots$) is transcribed as a decision variable alongside the control inputs. To enforce continuity, the state at the end of segment $[t_i, t_{i+1}]$ is constrained to match the decided starting state of the next segment ($x_{i+1}$), with the error termed as the ``defect''~\cite{direct-transcription-methods}. Shooting methods depend on numerical simulation to determine these states. In collocation methods, it is instead ensured that points on the trajectory segment (a.k.a. collocation points), coupled with the corresponding control inputs, comply with the system dynamics~\cite{direct-collocation}. The defect is an affine function of these collocation points.  

Motion Primitive and Trajectory Library methods are commonly used in aerial~\cite{traj_lib_drone,primitive_drone} robots, legged~\cite{traj_lib_humanoid} robots, and in Task And Motion Planning methods~\cite{compositional_models}. For these methods, pre-computed libraries of ideal or parameterized feasible trajectories are created. Valid sequences of these trajectories can generate approximately valid motions in nonlinear systems. Solutions can be refined using the same transcription methods mentioned earlier. 
All of these popular methods require solving often highly nonconvex programs in dynamic or underactuated systems to enforce considerations of dynamic feasibility or continuity. We propose a method to simplify these considerations to linear ones instead.

In our approach, we learn representations of sets of trajectories, termed Polytopic Action Sets, instead of individual trajectories. Each element within this linearly constrained set represents a feasible trajectory over a fixed horizon. Evaluating an appropriate trajectory within a Polytopic Action Set through our framework of Polytopic Action-Set and Motion Planning (PAAMP) can be formulated as a Linear Program. Furthermore, finding feasible motion plans over long horizons, given an appropriate sequence of Polytopic Action Sets, constitutes a Linear Program. Consequently, the motion planning problem can be approximated as a Mixed Integer Linear Program concerned with finding the appropriate sequences.\looseness=-1

Works in Multi-Modal Motion Planning (MMMP)~\cite{TAMPsurvey,perez_kaelbling_MMMP,MMMP_Hauser} have explored planning problems where a ``mode'' represents a distinct set of constraints. MMMP solutions are composed of motions where each trajectory segment satisfies the constraints of an associated mode. In finding sequences of Polytopic Action Sets, we adopt MMMP methods, such as Mode Adjacency Graphs. The use of negative methods in mode adjacency graphs aids in creating appropriate sequences of Polytopic Action Sets, enhancing the exploration of solutions to the MILP.\looseness=-1

The resulting formulation can be viewed as an extension of the Shortest Paths Problem described in ``Shortest Paths Through Graphs of Convex Sets'' (GCS)~\cite{GCS}, where a symbolic sequence represents a series of convex sets. GCS methods have successfully addressed planning problems in both the state~\cite{GCS_state_space} and configuration spaces~\cite{GCS_configuration_space} of a robot. With PAAMP, GCS solvers can be utilized to find motion plans where more abstract, dynamic constraints must be met. However, we have developed a heuristic that guides the analysis of the Mode Adjacency Graph, leading to an efficient sequence-then-solve approach.

We attempt to solve the Constraint Satisfaction Problem in motion planning over a variable horizon, in the presence of arbitrary constraints, using learned representations of feasible trajectories. The main contributions of this paper in achieving this goal are as follows:
\vspace{-5pt}
\begin{list}{}{\leftmargin=0em \itemindent=5pt}
    \item[i.] Introduce a novel methodology that simplifies the dynamic feasibility satisfaction problem by limiting the search to a finite number of polytopes;
    \item[ii.] Utilize these Polytopic Action Sets to facilitate a Linear Programming approach to long-horizon motion planning;
    \item[iii.] Adapt principles from Multi-Modal Motion Planning (MMMP) to establish a Mode Adjacency Graph, which enhances the exploration of action sequences;
    \item[iv.] Demonstrate the improved computational efficiency and practical applicability of the proposed motion planning solutions in real-world robotics scenarios.
\end{list}

\vspace{-5pt}
\section{Short Horizon Planning using Polytopic Action Sets}

Fixed horizon trajectory optimization typically requires defining an initial state, a final state, and ensuring that the trajectory parameters remain within a specific set, often non-convex.
In this section, we propose methods to tackle a variant of the short, fixed horizon trajectory parameter constraint satisfaction problem. Here, the set of feasible trajectories can be represented as being linearly constrained or as a union of linearly constrained sets.

\textbf{Subproblem 1.} \textit{Determine parameters $\omega \in \mathcal{W} \subseteq \mathbb{R}^n$ that define a trajectory $x(\cdot, \omega): [0,T] \rightarrow \mathbb{R}^r$ such that $x(0, \omega) = x_0$ and $x(T, \omega) = x_f$. Here, $x_0$ and $x_f$ represent the initial and goal states, respectively, and $\mathcal{W}$ denotes the set of all parameter values that yield dynamically feasible trajectories over the interval $[0, T]$. It is assumed that the trajectory $x(t, \omega)$ is affine with respect to the parameters $\omega$, and that $\mathcal{W}$ occupies a non-zero volume in $\mathbb{R}^n$.}

\subsection{Planning as a Linear Program}

\begin{figure}[!b]
    \centering
    \includegraphics[width=0.6\columnwidth,trim={30pt 20pt 50pt 75pt},clip]{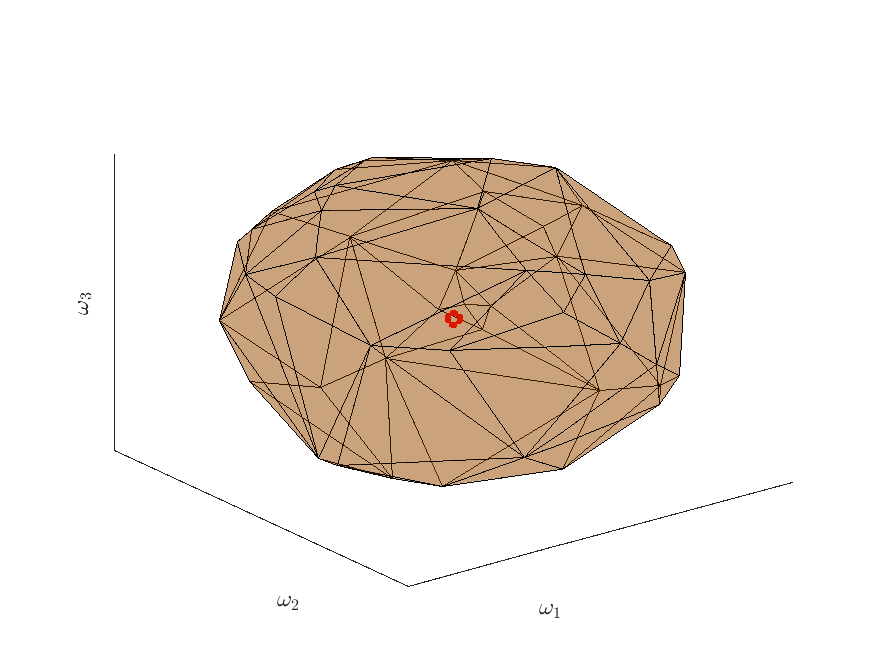}
    \caption{$\omega \in \mathcal{S}_w$ where $n = 3$.}
    \label{fig:ideal_W}
\end{figure}
\begin{figure}[!ht]
    \centering
    \includegraphics[width=\columnwidth,trim={30pt 65pt 14pt 80pt},clip]{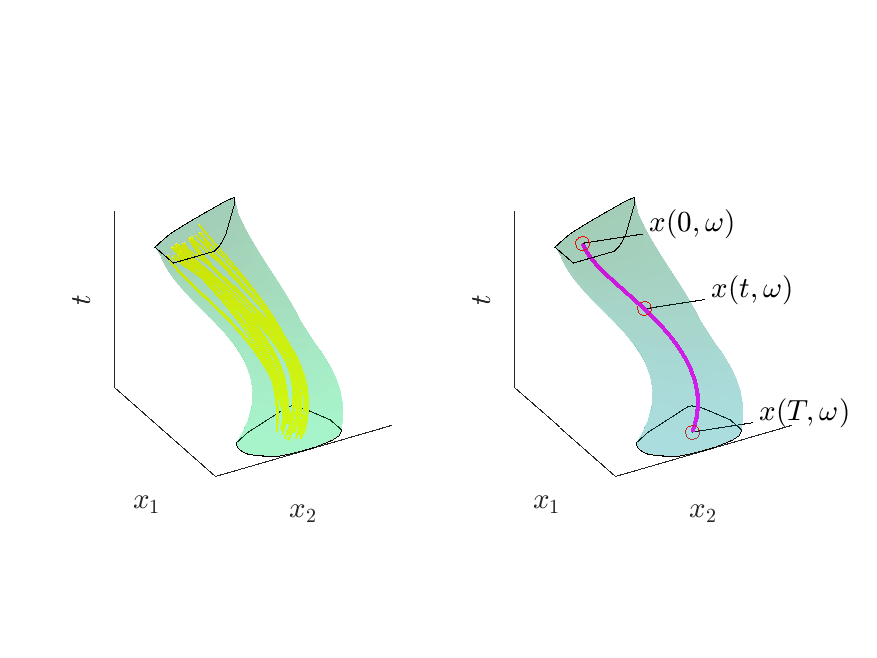}
    \caption{Example trajectories $x(\cdot, \omega)$, where  $\omega \in \mathcal{S}_w$.}
    \label{fig:funnel-1}
    \vspace{-10pt}
\end{figure}

We define a Polytopic Action Set as a linearly constrained set of parameter values, denoted as:
\begin{equation}
    \mathcal{S}_i = \{\omega \in \mathbb{R}^n \;|\; A_i \,\omega \leq b_i\} 
\end{equation}
with $A_i \in \mathbb{R}^{q_i \times n}$ and $b_i \in \mathbb{R}^{q_i}$. We enforce that $\mathcal{S}_i \subseteq \mathcal{W}$. Since $\mathcal{S}_i$ is a subset of $\mathcal{W}$, every parameter value within $\mathcal{S}_i$ parameterizes a valid trajectory.

The assumption that $x(t, \omega)$ is affine in $\omega$ leads to the following formulation:
\begin{equation}
    x(t, \omega) = H(t)\,\omega
\end{equation}
where $H(t)$ is a matrix-valued function of dimension $n \times r$, representing $x(t, \omega) \in \mathbb{R}^{r}$. Consequently, the boundary conditions, i.e., constraints on the initial and goal states, can be succinctly expressed as $H(0) \omega = x_0$ and $H(T) \omega = x_f$.

Consider searching for feasible parameter values within $\mathcal{S}_w$, a specific Polytopic Action Set, as shown in Fig.~\ref{fig:ideal_W}. In the context of Subproblem 1, we have $\mathcal{W} = \mathcal{S}_w$, and the problem of finding feasible parameters that meet the boundary constraints is cast as:
\begin{equation}\label{eq:subproblem1}
\begin{matrix}
\text{find any } \omega \in \mathbb{R}^n \;\text{s.t.} \\
A_w\,\omega \leq b_w, \\
\left[\begin{array}{c}
H(0) \\ 
H(T)
\end{array}\right] \omega = \left[\begin{array}{c}
x_0 \\ 
x_f
\end{array}\right]
\end{matrix}
\end{equation}
This formulation demonstrates that all constraints on $\omega$ are linear, thus defining a linear programming problem. 

It is feasible to incorporate additional linear constraints to this framework without compromising its linearity, such as constraints on intermediate states or bounds on specific parameters. Including convex, non-linear constraints transforms this into a convex, non-linear constraint satisfaction problem. However, our discussion remains focused on the linear case for simplicity.

Notably, $\mathcal{W}$ and $\mathcal{S}_w$ denote a set of parameters defining a trajectory within the fixed interval $[0,T]$ (as shown in Fig.~\ref{fig:funnel-1}), emphasizing that the solution $\omega \in \mathcal{W}$ is constrained to this fixed horizon.

\subsection{Short Horizon Planning as a MILP}\label{sebsec:short-MILP}

Consider now, searching for feasible parameter values over a union of $m$ Polytopic Action Sets, $\{\mathcal{S}_1, \mathcal{S}_2, ... \mathcal{S}_m\}$, where the union is given by $\mathcal{S}_W = \bigcup_{i=1}^m \mathcal{S}_i$. Given this definition, each parameter value $\omega$ in the set is associated with at least one Polytopic Action Set, $\mathcal{S}_i$, within the collection $\{\mathcal{S}_1, \ldots, \mathcal{S}_m\}$. This can be formally expressed as:
\begin{equation} \label{eq:w_is_in_one_subset}
    \forall\,\omega \in \mathcal{S}_W, \;\exists\; i \in \{1, \ldots, m\} : \;\omega \in \mathcal{S}_i
\end{equation}

To solve Subproblem 1 under these conditions ($\mathcal{W} = \mathcal{S_W}$), we reformulate the problem as a MILP. This formulation seeks a parameter vector $\omega \in \mathbb{R}^n$ and an integer $i \in \mathbb{Z}$, such that $\omega$ falls within $\mathcal{S}_i$ and satisfies the boundary conditions, leading to the following formulation:\looseness=-1
\begin{equation}\label{eq:subproblem1-new}
    \begin{matrix}
        \text{find any} \;\;\omega \in \mathbb{R}^n \;\;\text{and}\;\; i \in \{1, \ldots, m\} \;\;\text{s.t.} \\ 
        \omega \in \mathcal{S}_i, \\
        \left[\begin{array}{c}
        H(0) \\ 
        H(T)
        \end{array}\right] \omega = \left[\begin{array}{c}
        x_0 \\ 
        x_f
        \end{array}\right]
    \end{matrix}
\end{equation}

This problem formulation captures the concept of planning over short horizons as a MILP. Following~(\ref{eq:w_is_in_one_subset}), this approach aims to identify an appropriate $i$ and the corresponding Polytopic Action Set $\mathcal{S}_i$ that contains the solution $\omega$. In the worst-case scenario, this problem can be approached by solving $m$ linear sub-problems, derived from (\ref{eq:subproblem1-new}), examining each Polytopic Action Set ($\mathcal{S}_i$) in $\{\mathcal{S}_1, \ldots, \mathcal{S}_m\}$ for $\omega$.

\begin{figure}[!th]
    \centering    \includegraphics[width=0.7\columnwidth,trim={75pt 30pt 45pt 24pt},clip]{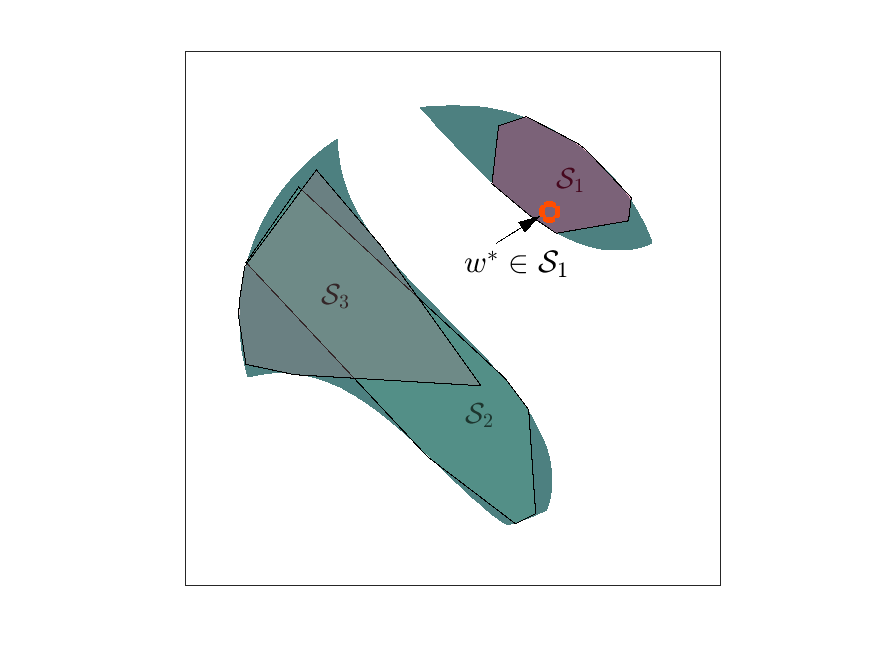}
    \caption{Decomposition of $\mathcal{W}$ with $n = 2$ and $m = 3$. Example solution, $\omega^*$, can be found using (\ref{eq:subproblem1-new}) with $\mathcal{S}_i = \mathcal{S}_1$.}
    \label{fig:decomposition}
    \vspace{-12pt}
\end{figure}

\subsection{Polytopic Approximations}
If $\mathcal{W}$, the set of all feasible parameter values for a system,  could be represented by a union of a finite number of Polytopic Action Sets, then the approach formulated by (\ref{eq:subproblem1-new}) would be complete. However, the practical realization that the set $\mathcal{W}$, though dependent on the specifics of the system, may not perfectly align with a union of polytopic sets, necessitates a method for approximation.

Notable strategies for achieving such an approximation include~\cite{ACD} and~\cite{IRIS}. Within the context of robotic C-Spaces, given kinematic constraints, strategies like the IRIS-NP method~\cite{IRIS_NP,werner2024approximating} are employed. Variations of this method facilitate the approximate decomposition of $\mathcal{W}$ into a series of polytopes, with the union denoted as $\mathcal{S}_W = \bigcup_{i = 1}^{m} \mathcal{S}_i$. $\mathcal{S}_W$ acts as an inner approximation to $\mathcal{W}$, but it is not an exact representation. Accordingly, while any parameter $\omega$ found within $\mathcal{S}_i$ guarantees feasibility, it does not ensure the completeness of the Constraint Satisfaction Problem (CSP). 

\section{Long Horizon Planning using Sequences of Polytopic Action Sets}

Building upon the foundational concepts of short horizon planning and the utilization of Polytopic Action Sets, we now extend our exploration to the domain of long horizon trajectory optimization. 

We maintain the assumption that a trajectory over a short horizon ($T$) is affine with respect to some parameter values $\omega$. A sequence of $l$ such trajectories, parameterized by $\omega_1, \omega_2, \ldots, \omega_l$, forms a longer horizon Piecewise Affine (PWA) trajectory, parameterized by $\hat{\omega} = [\omega_1, \omega_2, \ldots, \omega_l]^T$.
The longer horizon PWA trajectory is considered feasible if it is continuous and each segment lies within $\mathcal{W}$. For the longer horizon trajectory to be continuous, the condition $x(T, \omega_i) = x(0, \omega_{i+1})$ must be satisfied for all $i \in \{1, \ldots, l-1\}$.

In this section, we will simplify the discussion by equating the parameter values $\omega$ of a trajectory $x(\cdot, \omega)$ to the trajectory itself. This allows us to directly refer to a parameter set $\omega \in \mathcal{S}_i$ as a trajectory within $\mathcal{S}_i$. Consequently, the initial state of a trajectory will be denoted as $x_{initial}(\omega)$, and its conclusion as $x_{final}(\omega)$.

\textbf{Subproblem 2.} \textit{
Given the definitions of $\mathcal{W}$, $x_0$, and $x_f$ from Subproblem 1, find a vector of parameter values, $\hat{\omega} = \left[\omega_1, \, \cdots, \, \omega_l \right]^T$, that parameterizes a feasible long horizon trajectory. This trajectory must be continuous and satisfy the conditions $x_{initial}(\hat{\omega}) = x_0$, $x_{final}(\hat{\omega}) = x_f$, with $\omega_i \in \mathcal{W} \subseteq \mathbb{R}^n \; \forall i \in \{1, \ldots, l\}$. All trajectory segments are similarly defined and affine in parameter values.}

\subsection{Long Horizon Planning as a MILP}

The boundary constraints, ensuring the entire trajectory begins at $x_0$ and ends at $x_f$, remain linear. Continuing from~\ref{eq:subproblem1}, these constraints can be expressed as $H_b\,\hat{\omega} - x_b = 0$, where $x_b = \left[\begin{smallmatrix}x_0 \\ x_f\end{smallmatrix}\right]$ and $H_b$ captures the initial and final states of $\hat{\omega}$.

Similarly, the continuity constraint is linear in a PWA trajectory. For a consecutive pair of segments, $\omega_i$ and $\omega_{i+1}$, the continuity constraint can be encoded as $H(T) \omega_i - H(0) \omega_{i+1} = 0$. Consequently, we can represent the continuity constraint over the entire trajectory as $H_{c} \hat{\omega} = 0$. $H_{c}$ is constructed as an $r(l-1) \times nl$ sparse matrix that accounts for the continuity between each consecutive trajectory segment. 

Following this formulation, the fixed horizon planning problem where the number of segments, $l$, is predetermined, can be formalized as a non-linear program (NLP):
\begin{equation}\label{eq:5}
    \begin{aligned}
    \text{find any } & \hat{\omega} = 
    \left[\omega_1, \,\cdots,\,  \omega_l \right]^T
    \text{ subject to: } \\ 
    & \omega_i \in \mathcal{W}, \quad \forall i \in \{1, ..., l\}, \\
    & H_c \hat{\omega} = 0, \quad\;
    H_b \hat{\omega} - x_b = 0.
    \end{aligned}
\end{equation}

To navigate the complexity of the set $\mathcal{W}$, we employ an approximation through a union of Polytopic Action Sets (as depicted in Fig.~\ref{fig:decomposition}), still denoted as $\mathcal{S}_W$. As detailed in Section~\ref{sebsec:short-MILP}, each parameter vector $\omega_i$ within the approximated set $\mathcal{S}_W$ is contained in at least one polytope $\mathcal{S}_{\pi_i}$. Consequently, the original problem is transformed, enabling the identification of the specific Polytopic Action Set to which each $\omega_i \; \forall i \in \{1, ..., l\}$ belongs. This leads to a reformulation as a Mixed Integer Linear Program (MILP) as follows:
\begin{equation}\label{eq:sequence_program}
    \begin{aligned}
    \text{find any } & \hat{\omega} = 
    \left[\omega_1, \,\cdots,\,  \omega_l \right]^T
    \text{ and } 
    \pi \in \mathbb{Z}^l
    \text{ subject to: } \\ 
    & \omega_i \in \mathcal{S}_{\pi_i}, \quad \forall i \in \{1, ..., l\}, \\
    & 1 \leq \pi_i \leq m, \quad \forall i \in \{1, ..., l\},\\
    & H_c \hat{\omega} = 0, \quad\;
    H_b \hat{\omega} - x_b = 0.
    \end{aligned}
\end{equation}

Here, $\pi$ represents a sequence of integers corresponding to a sequence of Polytopic Action Sets. Given that this sequence is fixed, the program in (\ref{eq:sequence_program}) becomes a Linear Program. An admissible sequence $\pi$ is defined by the existence of a solution to the corresponding linear program.

Therefore, the motion planning problem is reduced to identifying an admissible sequence of Polytopic Action Sets.
In the worst case, every permutation of Polytopic Action Sets could be tested, which is not optimal. More efficient techniques for sequencing Polytopic Action Sets are preferable, and insights into these can be derived from the analysis of sequences of Polytopic Action Sets.

\begin{figure}[!th]
    \centering
    \includegraphics[width=0.75\columnwidth,trim={45pt 10pt 45pt 40pt},clip]{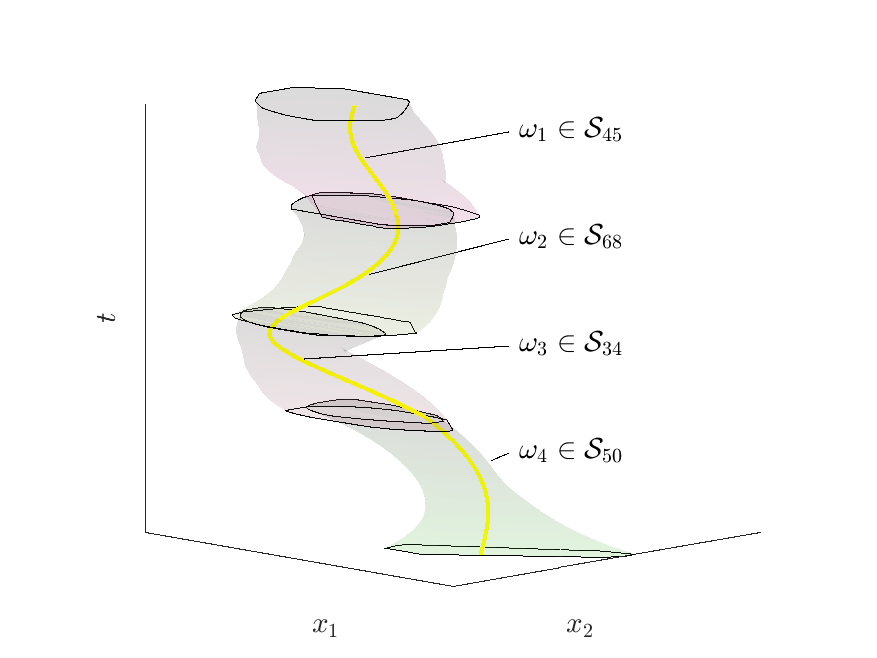}
    \caption{ A sample solution with sequence $\pi {=} \{45,68,34,50\}$.}
    \label{fig:solution_W3_with_forward_reachable_1}
    \vspace{-12pt}
\end{figure}

\subsection{Operations on Polytopic Action Sets}

\subsubsection{Backwards and Forwards Reachable Sets}

We define $\mathcal{X}_t : \mathcal{W} \rightarrow \mathbb{R}^r$ as a mapping that takes trajectories in $\mathcal{W}$ to their corresponding states at some time $t$.

Thus, $\mathcal{X}_0(\mathcal{S}_i)$ represents the Backwards Reachable Set of $\mathcal{S}_i$, which is the set of all initial states of a trajectory $\omega \in \mathcal{S}_i$. Formally, $
    \mathcal{X}_0(\mathcal{S}_i) = \{x_{initial}(\omega) \;\forall \omega \in \mathcal{S}_i\}$. Similarly, $
    \mathcal{X}_f(\mathcal{S}_i) = \{x_{final}(\omega) \;\forall \omega \in \mathcal{S}_i\}$
represents the Forward Reachable States, the set of all potential ``final'' states that can be achieved from executing any trajectory $\omega$ within $\mathcal{S}_i$. 

It naturally follows that if a certain state, $x_{pre}$ (resp. $x_{post}$) is not in $\mathcal{X}_0(\mathcal{S}_i)$ (resp. $\mathcal{X}_f(\mathcal{S}_i)$), then no trajectory exists with $x_{pre}$ as its initial state (resp. $x_{post}$ as its final state). 

\subsubsection{Forwards Reachable Sets Conditional to Initial States}
  
\begin{figure}[!th]
    \centering
    \includegraphics[width=0.625\columnwidth,trim={75pt 15pt 75pt 42pt},clip]{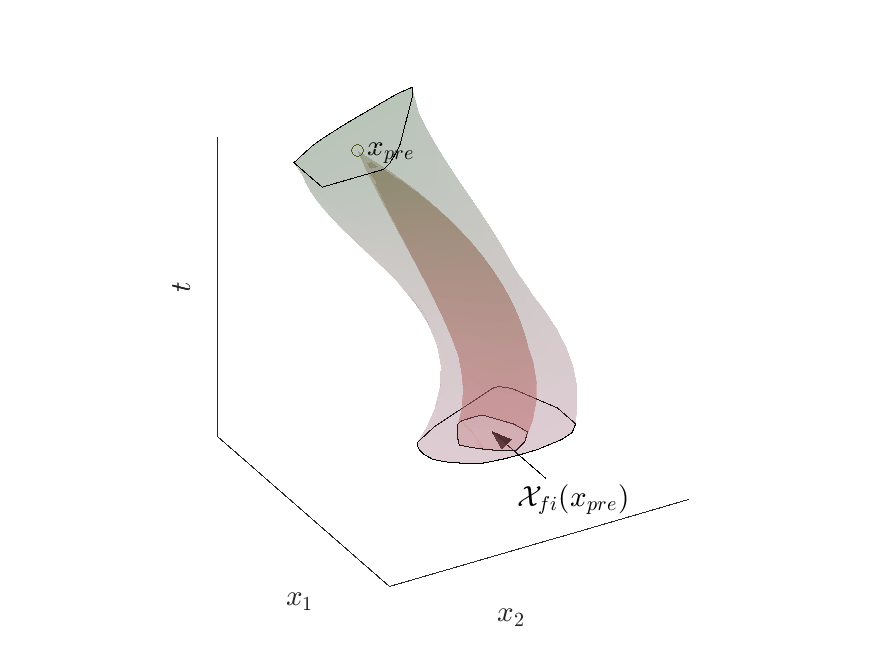}
    \caption{Visualizing $\mathcal{X}_{fi}(x_{pre})$.}
    \label{fig:FR_1}
    \vspace{-15pt}
\end{figure}

The concept of Forward Reachable States offers a comprehensive view of all possible final states achievable from a Polytopic Action Set. We now focus on the final states achievable from a specific initial state $x_{pre}$, as shown in Fig.~\ref{fig:FR_1}. This specificity is necessary for more nuanced planning.

For this purpose, we define $\mathcal{X}_{fi} : \mathcal{X}_0(\mathcal{S}_i) \xrightarrow{} \mathcal{X}_f(\mathcal{S}_i)$, which maps initial states to a subset of all Forward Reachable states, ensuring that the initial condition is also respected.
\begin{equation} \label{eq:Conditional_Forward_Reachable_Definition}
    \mathcal{X}_{fi}(\mathcal{X}) = \{x_{final}(\omega) \;|\; \omega \in \mathcal{S}_i \text{ and } x_{initial}(\omega)  = x_{pre} \in \mathcal{X}\}
\end{equation}

This definition captures the final states reachable through trajectories in $\mathcal{S}_i$ that specifically start from some $x_{pre} \in \mathcal{X}$. This mapping is generally non-injective, indicating the possibility of multiple $\omega \in \mathcal{S}_i$ corresponding to the same $x_{pre}$ but leading to different $x_{post} = x_{final}(\omega)$ values. Within PAAMP, the Forward Reachable Set of a Polytopic Action Set, conditional on a polytopic set of initial constraints, will always be polytopic.

Any value $x$ not in $\mathcal{X}_0(\mathcal{S}_i)$, the pre-image of $\mathcal{X}_{fi}$, has no corresponding trajectories $\omega \in \mathcal{S}_i$ and thus the associated Forward Reachable Set $\mathcal{X}_{fi}(x)$ is empty. Consequently, 
$\mathcal{X}_{fi}(\mathcal{X})$ is equivalent to $\mathcal{X}_{fi}(\mathcal{X} \cap \mathcal{X}_0(\mathcal{S}_i))$.

\subsubsection{Products\,of\,Polytopic\;Action\;Sets}
Finally we introduce the concept of a product of Polytopic Action Sets conditional to continuity constraints, denoted as $\mathcal{S}_i {\times_c} \mathcal{S}_j$. This product is defined to capture all feasible concatenated trajectories that transition smoothly from one action set to the next, adhering to the continuity constraints. Formally, it is defined as:\looseness=-1
\begin{align} \label{eq:Conditional_Product_Of_Set }
    \mathcal{S}_i \times_c \mathcal{S}_j = \hat{\mathcal{S}}_{i,j} &= 
    \left\{\hat{\omega} = 
    \left[\begin{matrix}
        \omega_1 \\ \omega_2
    \end{matrix}\right]
    \;\middle|\; \omega_1 \in \mathcal{S}_i, \omega_2 \in \mathcal{S}_j, H_c \hat{\omega} {=} 0 
    \right\} \nonumber \\
    &= \left\{\hat{\omega} \in \mathcal{S}_i \times \mathcal{S}_j \;\middle|\; H_c \hat{\omega} = 0 \right\}
\end{align}

The set $\mathcal{S}_i \times_c \mathcal{S}_j$, therefore, represents all feasible trajectories under a sequence $\pi = \{i,j\}$, where the individual trajectories from $\mathcal{S}_i$ and $\mathcal{S}_j$ are seamlessly connected by the continuity constraints at some state $x_c$. This state is necessarily forward reachable by $\mathcal{S}_i$ and backwards reachable by $\mathcal{S}_j$, so $x_c \in \mathcal{X}_0(\mathcal{S}_j) \cap \mathcal{X}_f(\mathcal{S}_i)$. If this intersection is empty, then there exists no continuous trajectories in $\mathcal{S}_i \times_c \mathcal{S}_j$, and the product would be an empty set. 

\begin{figure}[!th]
    \centering
    \includegraphics[width=0.65\columnwidth,trim={45pt 10pt 65pt 40pt},clip]{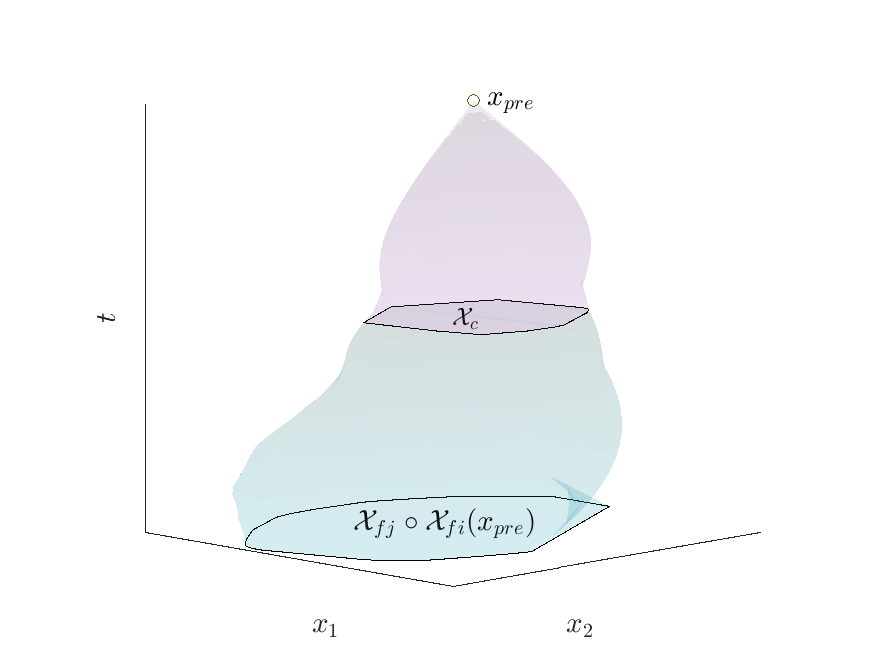}
    \caption{ 
        Trajectories in $\hat{\mathcal{S}}_{i,j}$ conditional on $x_{initial}(\omega) = x_{pre}$. 
    }
    \label{fig:FR_of_product}
    \vspace{-12pt}
\end{figure}

The existence of an intermediate state $x_c$ implies the existence of feasible concatenated trajectories and associated forward reachable states, $\mathcal{X}_{fj}(x_c)$. When extended to sets of forward reachable $x_c$ values, the Forward Reachable Set of $\mathcal{S}_{i,j}$ becomes $\mathcal{X}_{fj}( \mathcal{X}_{fi} ( \mathcal{X} ) )$, as shown in Fig.~\ref{fig:FR_of_product}. 

The operations defined for individual Polytopic Action Sets can similarly be extended to products of these sets. When considering longer sequences of actions, the product of two concatenated sequences of Polytopic Action Sets, $\hat{\mathcal{S}}_{\pi1} {\times_c} \hat{\mathcal{S}}_{\pi2}$, represents the combined set of actions for this extended sequence.
In this context, $\mathcal{X}_{f\pi}(\mathcal{X}) {=} \mathcal{X}_{f\pi_l}( ... \mathcal{X}_{f\pi_2}(\mathcal{X}_{f\pi_1}(\mathcal{X})) ... ) {=} \mathcal{X}_{f\pi_l}\circ \mathcal{X}_{f\pi_{l-1}} ... \mathcal{X}_{f\pi_2} \circ \mathcal{X}_{f\pi_1}(\mathcal{X})$ represents the Forward Reachable Set of the combined sequence $\hat{\mathcal{S}}_{\pi}$, conditional on the initial state being within the set $\mathcal{X}$.
This operation enables the analysis of feasible final states that are reachable from a given set of initial states, considering the entire sequence of actions encoded in $\hat{\mathcal{S}}_{\pi}$.\looseness=-1

A sequence $\pi$ is deemed admissible if and only if the goal state, $x_f$, is forward reachable by $\hat{\mathcal{S}_\pi}$ conditional to initial state $x_0$. Therefore, analyzing appended Polytopic Action Sets through the resultant product ($\hat{\mathcal{S}_\pi}$) can serve as a key determinant in selecting suitable action sequences for long horizon trajectory planning.

\section{Identifying Admissible Action Sequences}

The task of solving the Constraint Satisfaction Problem (CSP), given a fixed symbolic admissible sequence (as specified by the fixed value of $\pi$ in \ref{eq:sequence_program}) can be formulated as a linear program. This section aims to find such an admissible sequence of Polytopic Action Sets.

\textbf{Subproblem 3.} \textit{
    Given an initial state $x_0$, a goal state $x_f$, and a set of Polytopic Action Sets $\{\mathcal{S}_1, \ldots, \mathcal{S}_m\}$, the objective is to identify a sequence of indices $\pi$ that define a corresponding sequence of Polytopic Action Sets $\mathcal{S}_{\pi_1}, \mathcal{S}_{\pi_2}, \ldots, \mathcal{S}_{\pi_l}$, such that the goal state $x_f$ is included in the Forward Reachable Set of the concatenated sequence of action sets, starting from $x_0$. Formally, the goal is to ensure
    $$x_f \in \mathcal{X}_{f\pi_l} \circ \mathcal{X}_{f\pi_{l-1}} \ldots \circ \mathcal{X}_{f\pi_1}(x_0).$$
}

This subproblem effectively translates the task of long-horizon planning into a sequence discovery problem. The challenge is in selecting the appropriate series of Polytopic Action Sets that, collectively, enable a feasible path from the initial state to the desired goal state. Any admissible sequence will invariably produce a trajectory that adheres to the system's dynamics and constraints.

\subsection{Adapting\;MMMP\;Principles\;for\;Action\;Sequence\;Planning}

Building on the idea that finding sequences of action sets is similar to solving a Multi-Modal Motion Planning (MMMP) problem, this section explores strategies to address Subproblem 3. In this context, the terms Polytopic Action Set and Mode are used interchangeably, as both concepts involve satisfying specific constraints within a given timeframe ($\omega_i$ must belong to $\mathcal{S}_{\pi_i}$). The objective is to solve this MMMP-like Constraint Satisfaction Problem (CSP) using a sequence-then-satisfy approach, as outlined in~\cite{TAMPsurvey,guo2023recent}. 

This method involves an initial symbolic search to identify viable sequences of symbolic states (modes), followed by efforts to find corresponding continuous parameter values. If a continuous solution is not feasible, the process iterates with a new symbolic sequence.
This is dissimilar to some MILP solving methods, such as branch-and-bound~\cite{LanDoi60}.

To generate promising symbolic sequences, we propose the construction of a directed Mode Adjacency Graph ($G$). Each vertex in $G$ corresponds to a Polytopic Action Set, while the edges represent potential transitions or switching actions between these sets. An edge between vertices $\{\pi_i, \pi_{i+1}\}$ signifies a valid switching action $a(\pi_{i},\pi_{i+1})$, transitioning from some $\omega_{i} \in \mathcal{S}_{\pi_i}$ to $\omega_{i+i} \in \mathcal{S}_{\pi_{i+1}}$. A path through $G$ thus forms a ``skeleton'' or a ``candidate'' sequence for testing its admissibility, and can be identified using a K-Shortest Graph Search algorithm~\cite{k_shortest}. 

\begin{algorithm}\label{alg:seq-solve}
\caption{Sequence-Then-Solve}
\hspace*{\algorithmicindent} \textbf{Input} \; $\mathcal{S}_W,\,G,\,x_0,\,x_f$ \\
\hspace*{\algorithmicindent} \textbf{Output} \; A valid path from $x_0$ to $x_f$
\begin{algorithmic}
\State $G_{full} \gets FindFullGraph(S_W, G, x_0, x_f)$
\State $PathGenerator \gets InitKShortest(G_{full})$
 \Do
    \State $\pi_{candidate} \gets PathGenerator.PopNextShortest()$
\doWhile{$not(isAdmissible(\pi_{candidate}))$}
\State \Return $find\_\hat{\omega}(S_W, \pi_{candidate}, x_0, x_f)$
\end{algorithmic}
\end{algorithm}

To manage the complexity of the Mode Adjacency Graph ($G$) and minimize the consideration of inadmissible sequences, we adopt a ``negative method'' focused on pruning edges that inevitably lead to inadmissible sequences. This pruning is based on the fundamental requirement that a switching action must yield a non-empty set $\mathcal{X}_f(\hat{\mathcal{S}}_{\pi})$. 

For appending a mode $j$ onto a sequence $\pi$, the resulting Forward Reachable Set $\mathcal{X}_{fj} \circ (\mathcal{X}_{f\pi_l} \circ \ldots \circ \mathcal{X}_{f\pi_1}(x_0))$ must not be empty for the sequence to be admissible, serving as the ``pre-condition'' for the feasibility of switching modes. While this precondition can be computed during runtime, the concept of ``composability'' offers a pre-emptive pruning rule. Specifically, if $\mathcal{S}_{i} \times_c \mathcal{S}_{j}$ is empty for any pair $\{i, j\}$, then $\mathcal{X}_{fj} \circ \mathcal{X}_{fi}(\mathcal{X})$ is empty for every $\mathcal{X}$ and the precondition for the switching action $a(i,j)$ can never be met. Consequently, the edge $i \rightarrow j$ in $G$ can be pruned. 

It is important to recognize that while every admissible sequence corresponds to a path through $G$, not all paths through $G$ are admissible. This discrepancy underlines the necessity of the sequence-then-satisfy approach shown in Algorithm~1, where potential sequences identified in $G$ are subsequently validated for their feasibility in satisfying the overall planning constraints.

\begin{figure}
    \centering
    \resizebox{0.95\columnwidth}{!}{%
    \begin{tikzpicture}[shorten >=1pt, node distance=2cm,auto,
  inner/.style={draw,dashed,fill=blue!5,thick,inner sep=3pt,minimum width=8em},
  outer/.style={draw=gray,dashed,fill=black!2,thick,inner sep=5pt}
  ]
    \node[state,fill=black!10] (x0) {$x_0$};
    \node[state,fill=blue!15] (s3) [above right of=x0,xshift=8pt] {$\mathcal{S}_3$};
    \node[state,fill=blue!15] (s5) [right of=s3] {$\mathcal{S}_5$};
    \node[state,fill=blue!15] (s2) [right of=s5] {$\mathcal{S}_2$};
    \node[state,fill=blue!15] (s7) [below of=s5,node distance=1.4142cm] {$\mathcal{S}_7$};
    \node[state,fill=black!10] (xf) [right of=s7,node distance=3.75cm] {$x_f$};
    \node[state,fill=blue!15] (s6) [below of=s2,node distance=2.8284cm] {$\mathcal{S}_6$};
    \node[state,fill=blue!15] (s1) [below of=s3,node distance=2.8284cm] {$\mathcal{S}_1$};
    \path[->]
    (x0) edge (s3)
    (s5) edge (s3)
    (s5) edge (s2)
    (s2) edge (xf)
    (s6) edge (xf)
    (s1) edge (s6)
    (x0) edge (s1)
    (s7) edge (s5)
    (s3) edge (s7)
    (s1) edge (s7)
    (s2) edge (s6)
    (s6) edge[bend right] (s2)
    (s6) edge[out=-150,in=-30] node [name=y2]{}  (s1)
    (s5) edge [loop above] node [name=y1,yshift=-5pt]{} (s5);
    \begin{pgfonlayer}{background}
    \node[outer,fit=(x0) (xf) (s2) (s1) (y1) (y2)] (A) {} node [xshift=1pt,yshift=66pt] {$\bm{G_{full}}$};
    \node[inner,fit=(s2) (s1) (y1) (y2)] (A) {} node [xshift=42pt,yshift=66pt] {$\bm{G}$};
    \end{pgfonlayer}
    \end{tikzpicture}}
    \caption{Graph with candidate sequence:\,$\{\mathcal{S}_3,\mathcal{S}_7, \mathcal{S}_5,\mathcal{S}_5,\mathcal{S}_2\}$.}
    \label{fig:sequence}
    \vspace{-12pt}
\end{figure}
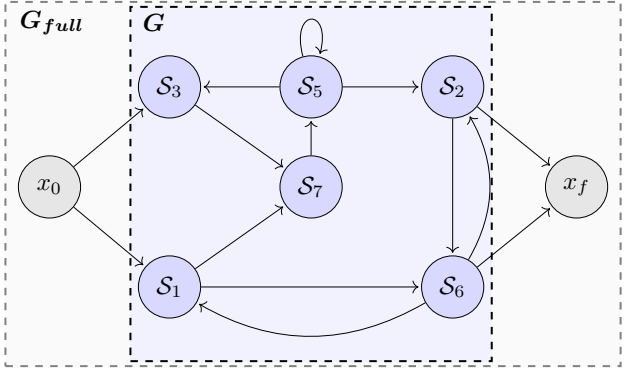

We add vertices representing the boundary conditions ($v_{x_0}, v_{x_f}$) to $G$ to form a new graph, $G_{full}$, also referred to as the Full Mode Adjacency Graph. A viable path through $G_{full}$ is characterized by the form $\{v_{x_0}, \pi_1, \ldots, \pi_l, v_{x_f}\}$, where $\pi = \{\pi_1, \ldots, \pi_l\}$ constitutes a candidate sequence as shown in Fig. \ref{fig:sequence}.

If $x_f \notin \mathcal{X}_f(\mathcal{S}_{\pi_l})$ or $x_0 \notin \mathcal{X}_{0}(\mathcal{S}_{\pi_1})$, then $x_f \in \mathcal{X}_{f\pi_l} \circ \ldots \circ \mathcal{X}_{f\pi_1}(x_0)$ cannot be satisfied. So, any edge leading from $v_{x_0}$ to a vertex $j$ in $G_{full}$ can be pruned if $x_0 \notin \mathcal{X}_0(\mathcal{S}_{j})$. Edges to $v_{x_f}$ from a vertex $j$ can be pruned if $x_f \notin \mathcal{X}_f(\mathcal{S}_{j})$.

\vspace{-1pt}
\subsection{Volume-Based Heuristics for Sequence Selection}\label{sebsec:volume_based_heuristic}
\vspace{-2pt}
In the quest to identify candidate sequences with a high likelihood of admissibility, we focus on the volume of the product of Polytopic Action Sets, $\hat{\mathcal{S}}_\pi = \mathcal{S}_{\pi_1} \times_c \ldots \times_c \mathcal{S}_{\pi_l}$. A larger volume of $\hat{\mathcal{S}}_\pi$ generally suggests a broader range of feasible solutions, implying a higher probability that the sequence $\pi$ is admissible. This inference is based on the premise that a larger $\hat{\mathcal{S}}_\pi$ correlates with expansive volumes of both $\mathcal{X}_0(\hat{\mathcal{S}}_\pi)$ and $\mathcal{X}_f(\hat{\mathcal{S}}_\pi)$, which are desirable traits for potential solutions.

To effectively leverage this insight in our k-shortest paths search, we propose a cost heuristic, $\ell(j,k)$, for each edge in the Mode Adjacency Graph $G$. The goal is to inversely correlate higher path costs ($\sum_{i = 1}^{l-1} \ell(\pi_i,\pi_{i+1})$) with lower likelihoods of path admissibility, as the K-Shortest Graph search will return low cost candidates first. 

The product of two Polytopic Action Sets, $\hat{\mathcal{S}}_{i,j}$, does not necessarily have a straightforward volume in $\mathbb{R}^{2n}$ due to the equality constraints, given by $H_c$. Instead, we represent the product as a new set, $\Lambda_{i,j}$, with a definable volume in $\mathbb{R}^{2n - r}$.

The set of inequalities defining $\Lambda_{i,j}$ can be expressed as:
\begin{equation} \label{eq:Lambda_i_j}
    \Lambda_{i,j} {=} \left\{\lambda \in \mathbb{R}^{2n - r} \,\middle|\,
    \left[
    \begin{matrix}
        A_i & 0 \\ 0 & A_j 
    \end{matrix}\right]
    Null(H_c) \lambda
    - \left[\begin{matrix}
        b_i \\ b_j
    \end{matrix}\right] \leq 0\right\}
\end{equation}
$Null(H_c)$ is a matrix representing the kernel of $H_c$, such that $Null(H_c)\lambda = \hat{\omega}$, where $H_c \hat{\omega} = 0$.

This set $\Lambda_{i,j}$ thus defines a new polytopic volume, indicative of the size of the original set $\hat{\mathcal{S}}_{i,j}$. By designing the edge cost $\ell(i,j)$ to be inversely related to the volume of $\Lambda_{i,j}$, the summation $\sum_{i = 1}^{l-1} \ell(\pi_i, \pi_{i+1})$ across a path $\pi$ becomes inversely correlated to the combined volumes of each successive pair in the sequence $\pi$. This is not exactly equivalent to the volume of the entire sequence, but it is well correlated. Consequently, this heuristic guides the k-shortest paths search towards those sequences that are not only feasible but also offer a broad range of solutions within the constraints of the problem.

\section{Experiments and Results}

\subsection{Experimental Setup}

Our experiment examines the motion planning of a torque-limited pendulum system, focusing on achieving a swing-up motion from an initial state of stable equilibrium to a desired final state. The pendulum system is characterized by its angle from stable equilibrium ($q$) and angular velocity ($\dot{q}$), which together define the state $x=\left[\begin{smallmatrix} q\\\dot q \end{smallmatrix}\right]$. The objective is to transition from an initial state $x_0 = [0,0]^T$ to a final state $x = [\pi,0]^T$, representing a complete swing-up motion.

\begin{figure}[!th]
    \centering
    \includegraphics[width=0.5\columnwidth,trim={0pt 0 0pt 0pt},clip]{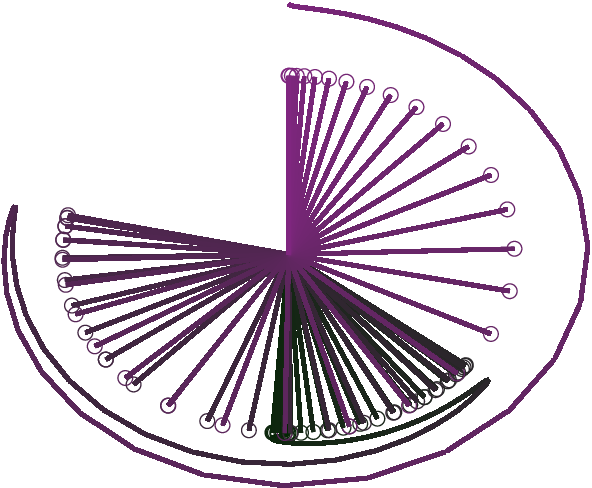}
    \caption{A possible swing-up trajectory solution.}
    \label{fig:swing-up}
    \vspace{-12pt}
\end{figure}

The pendulum has a mass of 0.1 kg ($m=0.1$) and is positioned 1 meter ($l=1$) from the pivot, with the rest of the pendulum assumed to be massless. Due to torque limitations, achieving a swing-up motion through a simple, monotonic increase in $q$ is not feasible. Instead, the strategy involves ``energy shaping'', where the pendulum is oscillated back and forth to incrementally gather the energy necessary to reach the goal state.
A representative motion of the pendulum swinging up is shown in Fig.~\ref{fig:swing-up}.

The trajectory of the angular velocity is defined using a Bezier curve, parameterized by 
$n-1$ coefficients. The initial angular position, combined with these $n-1$ parameters for angular velocity, comprehensively specifies a trajectory within the state space. Given that the Bezier curve is essentially a polynomial defined by these parameters, the state trajectory is affine in $\omega$, allowing for a linear programming approach to motion planning.

\vspace{-10pt}

\subsection{Decomposing Feasible Trajectories with Discrete Torque Constraints}

To address the challenge of enforcing the torque constraint
$|u(t,\omega)| \leq u_{max} \,\forall t \in [0, T]$
over a continuous interval, we opt for a discrete enforcement approach. Specifically, the constraint is applied at a set of discrete time points $|u(t,\omega)| \leq u_{max} \,\forall t \in \{0, t_1, t_2, ..., t_{d-1}, T\}$.

Our approach to decomposing the set of feasible trajectories employs a variant of the IRIS-NP algorithm. Unlike the standard method, which relies on randomized searches for counter-examples, our adaptation seeks potential counter-examples by sampling the surface of an expanding hyper-sphere. This strategy aims to identify large feasible regions within the state space efficiently.
The success of the decomposition heavily relies on the strategic seeding of the IRIS-NP algorithm. A particularly effective seed is a value of $\omega$ that satisfies $u(t, \omega) = 0 \; \forall t \in [0, T]$, representing a trajectory of free fall. This choice is predicated on the assumption that such a trajectory lies comfortably within the interior of the feasible set $\mathcal{W}$ given a positive $u_{max}$. By fitting $\omega$ values to these free-fall trajectories, we generate initial seeds for the IRIS-NP algorithm, aiming to maximize the size of identified feasible regions.

\subsection{Generating the Mode Adjacency Graph with Volume-Based Heuristics}

In accordance with Section~\ref{sebsec:volume_based_heuristic}, we developed a heuristic $\ell(i,j)$ based on the volume of associated $\Lambda_{ij}$. To construct the heuristic for this experiment, we first determined the approximate volume of $\Lambda_{ij}$, $Vol(\Lambda_{ij})$, for each valid switching action. A normal distribution was fit to this data, with $\mu$ and $\sigma$ representing the mean and standard deviation, respectively. $\ell(i,j)$ is a linear mapping of $Vol(\Lambda_{ij})$, where values of $Vol(\Lambda_{i,j}) = \mu$ were mapped to $\ell(i,j) = \ell_{max}$.
Intersections with lower volumes were pruned. Intersections where $Vol(\Lambda_{i,j}) = \mu + k_{max} \sigma$ were mapped to $\ell(i,j) = \ell_{min}$. $\ell_{min}$, $\ell_{max}$, and $k_{max}$ are parameters decided on a-priori. Volumes greater than $\mu + k_{max} \sigma$ are considered at a cost of $\ell_{min}$ as well. This systematic approach allows for the derivation of an approximate heuristic cost, $\ell(i,j)$, based on the volume-driven probability of path admissibility.

\subsection{Efficiency and Efficacy of the Proposed Approach}

Our experimental framework involved altering key parameters -- namely, the number of coefficients $n$, the planning horizon $T$, and the maximum torque $u_{max}$ -- to generate eight distinct systems ($\mathcal{W}$) and their corresponding approximated decompositions ($\mathcal{S}_W$), as listed in Table~\ref{tab:system}. Note that systems 1, 3, 5, 7 represent the same pendulum with different decomposition parameters. For each experiment we varied $\ell_{max}$, while setting $\ell_{min} {=} 1$ and $k_{max} {=} 2$ across all experiments.

Key metrics recorded include the computational time required to solve for the swing-up motion ($t_{solve}$), the number of candidate sequences evaluated ($k$), and the temporal length of the solution trajectory ($lT$). The initialization phase, which includes the generation of the Full Mode Adjacency graph and the setup for the K-Shortest Graph Search, was uniformly executed in approximately 30 ms for each trial.

\begin{table}
    \centering
    \caption{\small Indexed systems being considered}
    \begin{tabular}{cccc}
        \toprule
        $\mathcal{W}$ \# & $n$ & $T$ & $u_{max}$\\  
        \toprule
        1 & 5 & 0.5 & 0.5 \\
        2 & 5 & 0.5 & 0.6 \\
        3 & 5 &  1  & 0.5 \\
        4 & 5 &  1  & 0.6 \\
        5 & 6 & 0.5 & 0.5 \\
        6 & 6 & 0.5 & 0.6 \\
        7 & 6 &  1  & 0.5 \\
        8 & 6 &  1  & 0.6 \\
        \bottomrule
    \end{tabular}
    \label{tab:system}
    \vspace{-10pt}
\end{table}

\begin{table}
    \centering
    \caption{\small Measured effectiveness of the approach for each system }
    \vspace{-5pt}
    \begin{tabular}{|c|c||c|c|c|}
        \hline
        $\mathcal{W}$ \# & $ \ell_{max}$ & $t_{solve}$ (ms) & $k$ & $lT$ (s)\\ 
        \hline
        1 & 9 & 454.718 & 498 & 3\\
        1 & 300 & 22.038 & 23 & 3\\
        
        \hline 
        
        2 &  9 & 147.198 & 262 & 2.5\\
        2 &  300 & 15.346 & 27 & 2.5\\
        
        \hline
        
        3 & 9 & 6.433 & 9 & 4\\
        3 & 300 & 3.636 & 5 & 4\\
        
        \hline
        
        4 & 9 & 17.322 & 50 & 4\\
        4 & 300 & 2.614 & 6 & 4\\
        
        \hline
        
        5 & 9 & 10.838  & 9 & 3\\
        5 & 300 & 5.013  & 2 & 3\\
        
        \hline
        
        6 & 9 & 2.598 & 2 & 2.5\\
        6 & 300 & 2.829 & 2 & 2.5\\
        
        \hline
        
        7 & 9 & 0.961 & 1 & 4\\
        7 & 300 & 1.302 & 1 & 5\\
        
        \hline
        
        8 & 9 & 0.743 & 1 & 3 \\
        8 & 300 & 1.333 & 1 & 5 \\
        \hline
    \end{tabular}
    \label{tab:effectiveness_of_approach}
    \vspace{-5pt}
\end{table}

\begin{table}
    \centering
    \caption{\small{Measured effectiveness\,of\,heuristic\,for\,difficult sys.}}
    \vspace{-5pt}
    \begin{tabular}{|c|c||c|c|c|}
        \hline
        $\mathcal{W}$ \# & $ \ell_{max}$ & $t_{solve}$ (ms) & $k$\\  
        \hline
        1 & 1 & 6096.54 & 6501 \\ 
        1 & 9 & 454.718 & 498 \\
        \hline
        3 & 1 & 20.379 & 24 \\
        3 & 9 & 6.433 & 9 \\
        \hline
        5 & 1 & 247.375 & 153 \\
        5 & 9 & 10.833 & 9 \\
        \hline
    \end{tabular}
    \label{tab:effectiveness_of_heuristic}
    \vspace{-5pt}
\end{table}

\begin{figure}[!th]
    \centering
    \includegraphics[width=0.725\columnwidth,trim={45pt 6pt 50pt 45pt},clip]{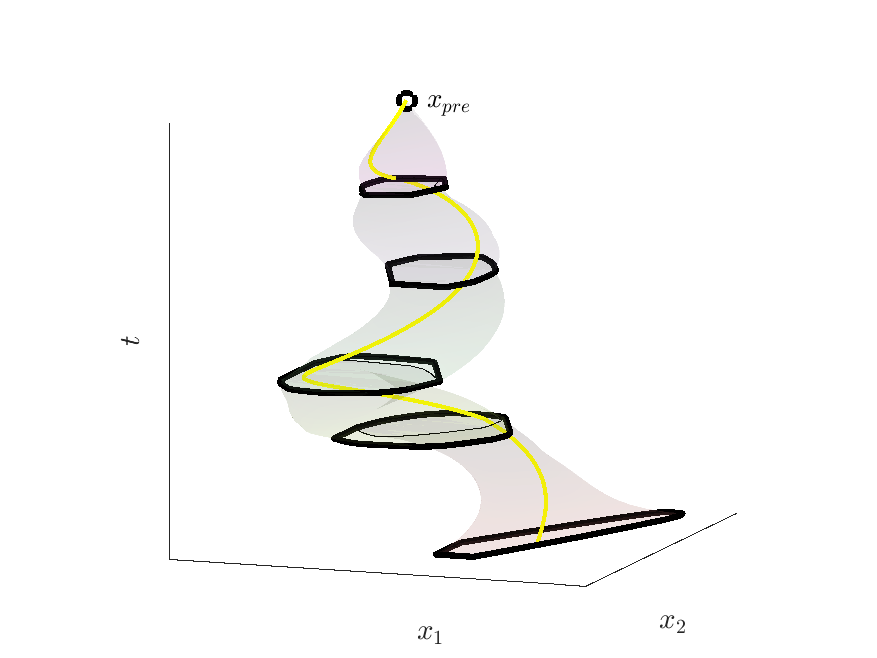}
    \caption{The states forward reachable from $x_{pre} = x_0$ through sequences $\{67\}, \{67, 35\}, ... \{67, 35, 88, 34, 50\}$ for $\mathcal{W}_3$.}
    \label{fig:solution_W3_with_forward_reachable_2}
    \vspace{-20pt}
\end{figure}

Our findings, as listed in Table~\ref{tab:effectiveness_of_approach}, highlight a direct correlation between the number of candidates tested and the computational efficiency of our approach. Specifically, in scenarios where a suitable action sequence is identified promptly ($k = 1$), our method achieves a resolution in less than 1 ms, as evidenced in our efficiency assessments. Conversely, when the process evaluates a larger pool of candidates ($k = 498$), the resolution time extends to nearly half a second.
This observation highlights the criticality of an appropriate heuristic for minimizing candidate evaluations in the proposed sequence-then-solve approach.

Adjusting the parameter $\ell_{max}$ effectively influences the likelihood of identifying an admissible sequence early in the search process. Our heuristic, which discriminates based on action set volume, plays a pivotal role here. By assigning a minimal cost to larger-volume actions, we can prioritize more promising action sequences. Notably, equalizing $\ell_{max} = \ell_{min} = 1$ negates the volume distinction, treating all actions uniformly.
Our experimental results, as presented in Table~\ref{tab:effectiveness_of_heuristic}, show that imposing a higher cost on actions with lower volumes (indicated by a greater $\ell_{max}$ value) speeds up the solution process, validating the efficacy of a volume-based heuristic. However, this strategy may inadvertently prioritize loops of inadmissible actions if they are assigned low costs, particularly if the loop recurs in candidate solutions, highlighting a potential area for refinement.
In the worst case, if a solution does not exist, a solver may inadvertently attempt to enumerate every path through the graph.

\subsection{Analyzing Solution Optimality and System Performance}

The method employed for solving the swing-up problem is framed as a Constraint Satisfaction Problem (CSP), where the focus is not on obtaining optimal solutions. The requirement for solutions to adhere to a horizon of $lT$ seconds also introduces inherent limitations on optimality, especially concerning time efficiency.

\begin{figure}[!t]
    \centering
    \includegraphics[width=0.76\columnwidth,trim={20pt 12pt 40pt 20pt},clip]{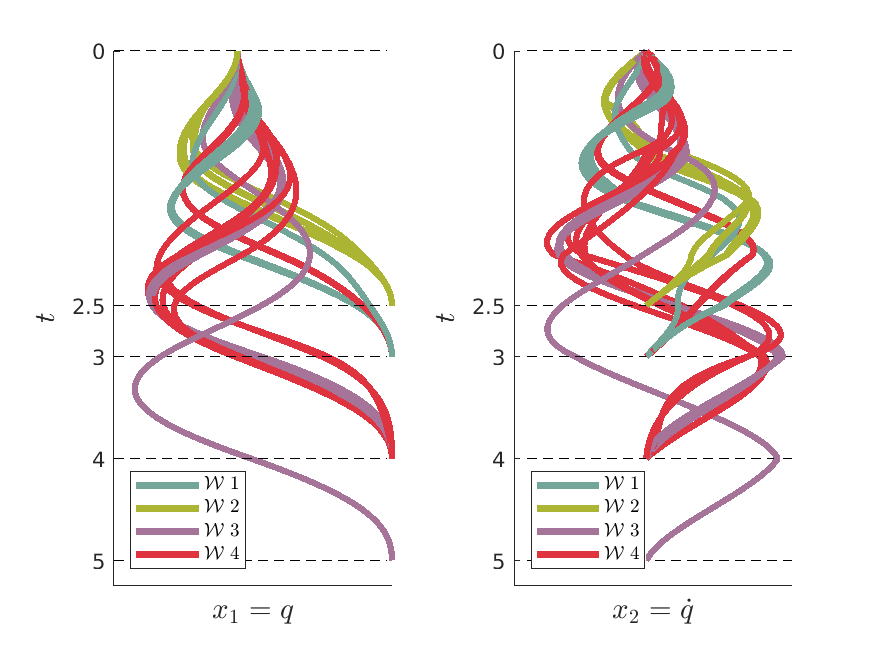}
    \caption{ Sample solutions for the first 4 Systems. Systems 1 and 2, with $T = 0.5$ solve for shorter trajectories ($lT = 2.5$s and $lT = 3$s). Systems 1 and 3 are more limited in torque, and thus generally need more swinging actions.}
    \label{fig:solutions_for_n_5}
    \vspace{-15pt}
\end{figure}

Systems configured with a shorter planning horizon ($T = 0.5$ seconds), such as $\mathcal{W}_1$ and $\mathcal{W}_2$, shown in Fig.~\ref{fig:solutions_for_n_5}, are more likely to produce solutions that are closer to time-optimal, with solution times spanning between 2.5 and 3 seconds, respectively.
However, these systems also exhibited longer computational times compared to their counterparts configured with $T = 1$ second. This observation can be attributed to the fact that, for a given total horizon, sequences requiring a shorter $T$ will inherently contain more steps, thereby prolonging the solution process.
This is because shorter sequences are prioritized by the K-Shortest Path algorithm, leading to increased computational effort.
Furthermore, systems with a higher number of parameters ($n=6$, specifically systems 5 through 8) demonstrated quicker solution times. Note that systems 5 and 1 represent the same pendulum, with only $n$ varied. This faster performance can be attributed to the increased diversity in the trajectories described by Polytopic Action Sets with a larger $n$, providing a broader spectrum of potential solutions for evaluation.

\noindent Additionally, our findings highlight the potential robustness of a solution. As illustrated in Fig.~\ref{fig:solution_W3_with_forward_reachable_2}, depicting solutions for $\mathcal{W}_3$, sequence $\pi$ is admissible for large regions of goal states, including those in the neighborhood of $x_f$. This principle also applies reciprocally for the backward reachability of $x_0$, where $\pi$ maintains admissibility given variations of $x_0$.

\vspace{-3pt}
\section{Conclusions and Future Work}
\vspace{-3pt}

In conclusion, this study presented a novel approach to solving long-horizon, dynamically feasible motion planning problems using Polytopic Action Sets, demonstrating the potential for efficient motion planning in robotic systems.
By leveraging learned representations known as Polytopic Action Sets, we have efficiently computed long-horizon trajectories and transformed these complex planning problems into sequences of linear programs.

Future work will explore alternative strategies for MMMP to generate candidate sequences, aiming for faster solutions. Considering that certain sequences $\pi$ are generally admissible around initial and final states, there is potential to develop a robust continuous motion planner or controller that uses previous solutions to inform future searches.

\vspace{-2pt}
\section{Acknowledgements}
\vspace{-2pt}
    Thanks to Ethan Chandler, Hushmand Esmaeli, and Ya\c{s}ar \.{I}dikut for key discussions on the concepts of composability, feasible sets, and appropriate notation. Thanks to Mark Petersen, Russ Tedrake and Rebbecca Jiang for conversations regarding decomposition, IRIS-NP, and GCS.

\bibliographystyle{IEEEtran}
\bibliography{bibtex}

\end{document}